\documentclass[a4paper,10pt]{llncs}

\usepackage{amssymb}
\usepackage{graphicx}
\usepackage{indentfirst}

\usepackage{tabularx,xspace} 
\usepackage{wrapfig,multirow,booktabs}

\usepackage{url}

\urldef{\mailsa}\path|laurynas.karazija@cantab.net,{petar.velickovic, pietro.lio}@cl.cam.ac.uk| 
\newcommand{\keywords}[1]{\par\addvspace\baselineskip
\noindent\keywordname\enspace\ignorespaces#1}

\makeatletter
\def\@seccntformat#1{\csname the#1\endcsname~}
\makeatother

\begin{document}

\mainmatter  

\title{Automatic Inference of Cross-modal Connection Topologies for X-CNNs}

\titlerunning{Automatic Inference of Cross-modal Connection Topologies for X-CNNs}

\def\petar{Veli{\v{c}}kovi{\'c}\xspace}
\author{Laurynas Karazija \and Petar \petar
\and Pietro Li\`o}

\authorrunning{Automatic Inference of Cross-modal Connection Topologies for X-CNNs}

\institute{Computer Laboratory, University of Cambridge, Cambridge, United Kingdom.\\ \url{laurynas.karazija@cantab.net,}\url{{pv273, pl219}@cam.ac.uk}}

\toctitle{Automatic Inference of Cross-modal Connection Topologies for X-CNNs}
\tocauthor{Authors' Instructions}
\maketitle

\begin{abstract}
This paper introduces a way to learn cross-modal convolutional neural network (X-CNN) architectures from a base convolutional network (CNN) and the training data to reduce the design cost and enable applying cross-modal networks in sparse data environments. 
Two approaches for building X-CNNs are presented. The base approach learns the topology in a data-driven manner, by using measurements performed on the base CNN and supplied data. The iterative approach performs further optimisation of the topology through a combined learning procedure, simultaneously learning the topology and training the network. The approaches were evaluated agains examples of hand-designed X-CNNs and their base variants, showing superior performance and, in some cases, gaining an additional 9\% of accuracy. From further considerations, we conclude that the presented methodology takes less time than any manual approach would, whilst also significantly reducing the design complexity. The application of the methods is fully automated and implemented in Xsertion library.\footnote{Code is publicly available at \url{https://github.com/karazijal/xsertion}.}
\end{abstract}

\section{Introduction}
In recent years, deep learning has become a popular approach, revolutionising various fields, including computer vision~\cite{krizhevsky2012imagenet}, agent control~\cite{mnih2015human} and natural language processing~\cite{hinton2012deep}.
However, training such deep neural nets requires vast amounts of labeled data, a limiting factor in many fields. An example of this is biomedical research, where few data examples can be obtained because the number of patients taking part in clinical studies is limited. 

A proposal by \petar et al.~\cite{velivckovic2016x} has introduced cross-modal convolutional networks (X-CNNs) that use several constituent CNNs to process parts of wide data. This architecture has shown performance improvements in sparse data environments~\cite{velickovic2017cross}, offering a new way to handle different types of data present (different \textit{modalities}).
In X-CNNs, separate CNN super-layers are utilised to process each modality individually, producing specialised feature detectors. The classifier part of the network is shared between super-layers. Additionally, \textit{cross-modal connections} are established between super-layers to share information between modalities. 
However, this architecture carries a significant overhead -- the number of design decisions concerning the network composition grows at least quadratically with the number of different types of data available. This means that such a technique is not well suited for widespread application. 

This paper proposes a solution to this problem by introducing two methods to construct cross-modal architectures automatically, learning the required architecture from a base untrained convolutional network and a portion of the training data. This serves to both reduce the design effort and facilitate easy application by non-expert practitioners. Furthermore, the evaluation of the approaches showed that not only are the automatically constructed models better than hand-constructed ones, but also the additional time and design effort requirements are superior to those of the manual approaches. The next section discusses related work, followed by explanation of the methods, evaluation and conclusions.

\section{Related Work}

The idea to learn neural network architectures is not new and there has been substantial previous work on this. Initially, this was primarily achieved through the use of evolutionary algorithms and meta-heuristic optimisation approaches such as particle swarm optimisation~\cite{yao1999evolving}, evolutionary programming net by Yao et al.~\cite{yao1997epnet}, neuroevolution of augmenting topologies by Stanley et al.~\cite{stanley2005real} as well as variants of neural trees by Zhang et al.~\cite{zhang1997evolutionary} and Chen et al.~\cite{chen2005time} to name a few. However, all these approaches were aimed at a single MLP problem, and involved constructing ensembles of models through time-consuming runs of cross-validation whilst still operating in low-dimensional problem spaces. 

Initial attempts at adjusting deep neural nets algorithmically were primarily focused on reducing the computational complexity. Molchanov et al.~\cite{molchanov2016pruning} used train-prune iterations to remove connections from CNN models. Similarly, Hu et al.~\cite{hu2016network} utilised a more data-driven approach to remove whole neurons. Wen et al.~\cite{wen2016learning} proposed structured sparsity learning as a way to regularise CNNs to produce a more compact form. However, all these approaches relied on having a trained CNN and aimed to simplify its structure by removing elements for computational- and energy- efficiency with a minimal reduction in performance. This project, instead, aims to adjust and add topological elements to the CNNs algorithmically to increase performance.

Work by Zoph et al.~\cite{zoph2017nas}, Real et al.~\cite{real2017large} and Baker et al.~\cite{baker2016designing} revisited ideas of evolutionary algorithms, applying them to deep learning by training an agent to design NNs using reinforcement learning. The approaches showed competitive results but search space could not include cutting-edge topological modifications, and came with extensive computational, data and time costs. Work presented here concentrates on introducing one such novel topology automatically and could be used in conjunction with their work to achieve state-of-the-art results in low-data availability environments.

\section{Methodology}
Two methods, the base and iterative approach, take as input an untrained CNN to use as a blueprint for the sort of network that would be appropriate for the dataset. Then, a portion of training data is used to infer and introduce two topological aspects of X-CNNs: separation into super-layers and introduction of cross-modal connections. The methods produce networks with similar number of parameters to the input CNN, maintaining similar computational complexity.

It should be noted that the separation of data into different modalities falls outside the scope of these approaches, and could be handled either through unsupervised methods such as clustering or done manually through domain 
knowledge. To carry out work here, we utilised known modalities for visual data of \textit{luminance} and \textit{chroma difference}, drawing inspiration from the human visual system.

\subsection{Base Approach}
\begin{figure}[b]
\begin{center}
	\includegraphics[trim=0mm 5mm 0mm 5mm,width=\textwidth]{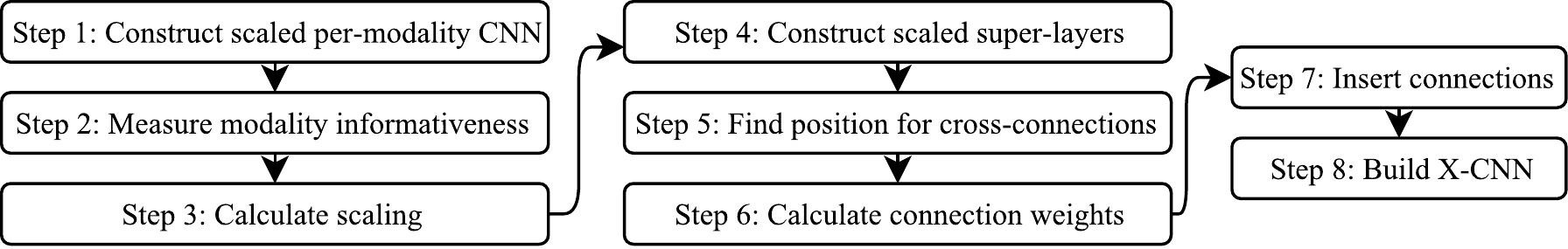}
\end{center}
	\caption{\small The steps of the base automatic approach to creating X-CNN topology.}
	\label{fig:base_approach_pipeline}
\end{figure}

We conducted a series of ablation experiments to investigate important aspects of X-CNN architecture. This allowed decomposing the problem of CNN\(\rightarrow\)X-CNN transform into several steps that can be carried out algorithmically.
In short, the base approach constructs the cross-modal architecture using measures of generalisation accuracy of each modality, to calculate hyper-parameters for each super-layer and connections between them, whilst a heuristically guided search is used to find connection positions. 
This reduces the design complexity from a quadratic of a number of modalities to two hyper-parameters \(\alpha, \beta\). The following details the key steps (Fig. \ref{fig:base_approach_pipeline}) in the base approach with relevant findings from the experiments.
\subsubsection{Modality Informativeness (Steps 1 and 2).}

To perform CNN \(\rightarrow\) X-CNN transformation some notion of how informative modality is was required. In deep learning, near-raw data is used in the models, so existing methodologies, such as feature selection and ranking, are not suited to examine the significance of modalities. Here, we found accuracy measures performed on the base CNN using only one modality to be suitable for the task, giving a modality informativeness measure \(n_{l_i}\) for modality \(i\).

\subsubsection{Super-layers (Steps 3 and 4).}

Super-layers are instantiated as copies of the base CNN, which take as input only one modality and share the classifier part of the network. However, experiments have shown that giving more feature maps to a more informative data split allows appropriate feature representation to be constructed whilst keeping the overall complexity of the network low. To support a variety of architectures and not be bound by dimensional constraints required by certain operations, a scaling multiplier, parametrised as follows, is used:
\begin{equation}
	s_{l_1} = \frac{n_{l_1}^\alpha}{\sum_i n_{l_i}^\alpha} .
	\label{eq:layer_scale}
\end{equation}
where \(\alpha\) is a hyper-parameter used to tune how much higher informativeness is prioritised. This multiplier scales appropriate hyper-parameters of layers within the super-layers, such as controlling number of kernels.

\subsubsection{Cross-modal Connections.}

The experiments have shown that cross-modal connections are key to capturing cross-modal interactions, without which model performance is severely degraded. 
The algorithm approaches building connections in three steps: firstly, finding a position where the connections could be placed, secondly, determining how super-layers should connect and, finally, determining the composition of the connection. 

\begin{figure}[t]
\begin{center}
\includegraphics[trim=0mm 5mm 0mm 5mm]{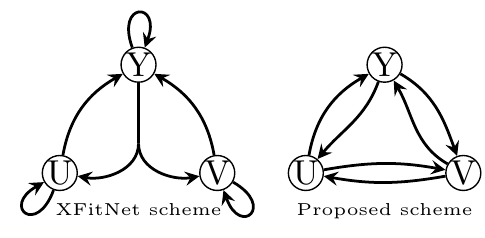}
\end{center}
	\caption{Problem of placing cross-modal connections shown as a directed graph.}
	\label{fig:dir_graph_nodes}
\end{figure}

\paragraph{Position (Step 5).} 
The place where cross-modal connections should be established in the network is more dependent on the CNN topology rather than data. 
We introduced a heuristic that places cross-modal connections at the ends of blocks or modules, such as commonly used downsampling operations following convolution~\cite{krizhevsky2012imagenet,romero2014fitnets} or various merge points used in other architectures~\cite{Szegedy_2015_CVPR,he2016deep,srivastava2015highway}. 
This approach was verified using linear probe\footnote{A linear classifier is trained using outputs of frozen intermediate layers as inputs, measuring generalisation performance.} methodology in~\cite{alain2016understanding}. Such points were shown to have a large ratio of accuracy to feature volume suggesting that a suitably dense representation exists that could be used as extra context by other modalities.
\paragraph{Connectivity (Step 7).}
The problem of placing cross-modal connections is modelled as a directed graph where nodes are points from super-layers and edges are connections (Fig. \ref{fig:dir_graph_nodes}). It is reasonable to assume that at the same depth feature extractors of similar complexities are formed.\footnote{For example, it is known that nearly always the initial convolutional layers in CNNs learn to be edge extractors~\cite{zeiler2014visualizing}.} We therefore always connect points at the same depth.  
This also allows circumventing the problems of projection such as those encountered by the authors of~\cite{he2016deep,srivastava2015highway}, which limited their ability to optimise the computational efficiency of the network. Additionally, since the connections are concatenated rather than summed, the network maintains the ability to utilise the information passed through the connections at lower depths if that is optimal.
 Experiments have shown that a fully-connected graph allows learning appropriate connections between each pair of modalities and sharing information between all.
 This leads to better performance than a more restricted variant used in XKerasNet and XFitNet~\cite{velivckovic2016x}.
\paragraph{Composition (Step 6).}
To abstract from specifics of operations performed on the connection, we introduced a concept of a connection \textit{weight}\footnote{This should not be confused with parameters of the layers themselves, in other literature sometimes referred to as weights as well.}, which controls hyper-parameters of these operations. 
The experiments showed that connections with equal weights did not perform as well as connections that were weighted more heavily for originating in a more informative node. This is because the connections implement a mapping accomplishing three things. Firstly, they \textit{compress} the information transferred along the connection through weighted combinations of outputs from the origin super-layer, reducing the parameter cost of destination super-layer. Secondly, they implement an \textit{affine transformation} of features. Finally, they provide \textit{gating} during training, which prevents gradients of a highly informative super-layer from propagating too much into a less informative one. Intermediate computations on the cross-connection soak up much of the transferred gradient to train the connection itself, enabling each super-layer to learn to be an optimal feature extractor for its respective modality.

The connection weight controls this behaviour. Thus, a desired weight \(w_{l_1,l_2} \in [0,1]\) of a connection between super-layers \(l_1 \rightsquigarrow l_2\) is such a number \(w_{l_1,l_2} > 0.5\) for connecting a node from a more informative position to a less informative one. A sensible parametrisation of this is 
\begin{equation}
	w_{l_1,l_2} = \frac{n_{l_1}^\beta}{n_{l_1}^\beta+n_{l_2}^\beta} ,
	\label{eq:con_weight}
\end{equation}
where \(\beta\) is a hyper-parameter controlling the discounting of lower informativeness. With \(\beta \rightarrow 0\) all nodes are treated equally. When \(\beta \rightarrow \infty\), most of the weight is assigned to the features transferred from the informative super-layer. If the weight is set to zero, the connection is dropped.

\subsection{Iterative Approach}

\begin{figure}
	\begin{center}
		\includegraphics[trim=0mm 5mm 0mm 5mm,width=\textwidth]{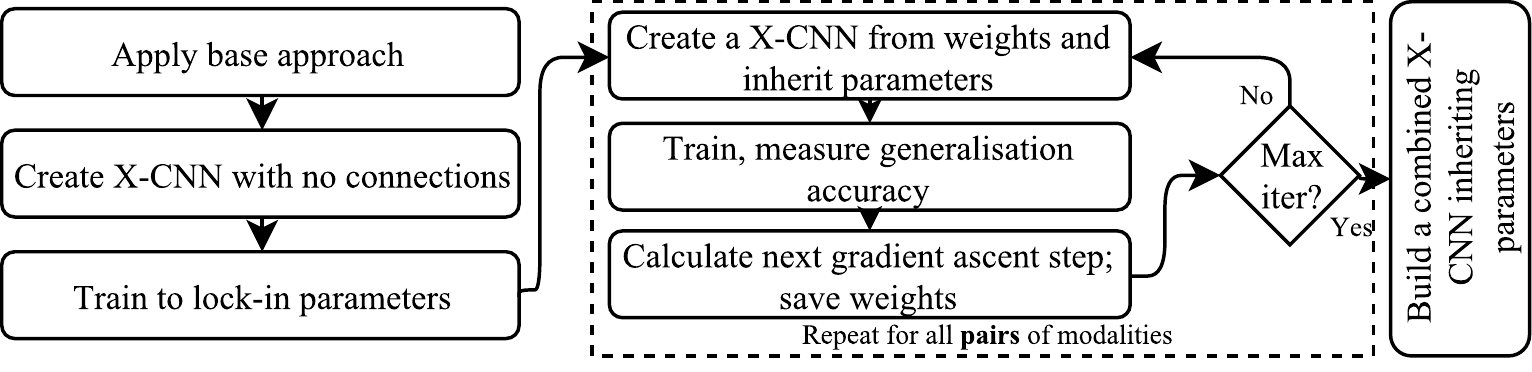}
	\end{center}
	\caption{Iterative approach for producing a trained X-CNN}
	\label{fig:iter_final}
\end{figure}

The iterative approach (Fig. \ref{fig:iter_final}) extends the previous methodology by adopting a learning procedure for the connection weights. 
It works by constructing successive generations of X-CNN models, where each generation contains models 
for each pair of modalities. The first generation uses equal-weighted connections, whilst second-generation uses the base approach calculations. Afterwards, a gradient ascent procedure on connection weights is performed. We adapt ideas behind Nasterov accelation for Adam optimiser~\cite{kingma2014adam,dozat2016incorporating} to construct an update procedure based on the generalisation measures for the connection weights. We also add weight-decay regularisation to ensure that resulting X-CNN does not grow too complex.

\paragraph{Parameter Inheritance.}
To make this approach viable in practice, additional technique is needed to reduce the time requirements.
An initial pre-training step is conducted where X-CNN without connections is trained for several epochs to lock parameter positions. Alternatively, the same random seed value can used for initialisation across models. This enables inheriting layer parameters between successive generations, making training of each generation require only a small number of updates to adapt to new connections, greatly reducing time requirement.

\paragraph{Combined Learning.}
The parameter inheritance transforms the gradient ascent procedure for connection-weights into a combined learning procedure for a full X-CNN, where both the connection-weights and model parameters are learned simultaneously. Effectively, the search is performed not in the parameter space of the resulting X-CNN, but across a combined parameter space of all potential X-CNNs with different connection weights. This is only possible because the search performed is inherently \textit{greedy}. It assumes that a better minima can be found using better connections weights rather than by continuing to optimise parameters. In other words, combined learning only takes steps on a subset of axes at once. This could lead to a problem where parameters become too finely tuned for a particular connection configuration, ``poisoning" future generations. To counteract this, we introduce a slight perturbation to the parameters between generations, by averaging across a couple of generations, which acts an additional regularisation against this type of overfitting.

\section{Evaluation}
The approaches were evaluated against hand-optimised X-CNNs from~\cite{velivckovic2016x}: XKerasNet based on KerasNet~\cite{chollet2015keras} and XFitNet based on Fitnet4~\cite{romero2014fitnets}. The datasets used were CIFAR-10/100~\cite{krizhevsky2009learning}. All models were trained using a learning rate of \(10^{-3}\) utilising Adam optimiser~\cite{kingma2014adam}. Training was done using batch size of 32/128 for 200/230 epochs for Keras/Fitnet based models. Xsertion library built X-CNN topologies using KerasNet and FitNet4 as blueprints, using 80\% of the training set for internal training for 80 epochs and 20\% of the training set as a validation set for internal metrics. Hyper-parameters \(\alpha\) of 1 and 2 and \(\beta\) of 2 and 4 were used, respectively. At each datapoint, \(p\)\% of \textit{per-class} samples were retained in the training set.

\subsection{Results}
\subsubsection{Base Approach.}
Tables \ref{tab:res_keras} and \ref{tab:res_fitnet} detail the results. The evaluation shows that models constructed using the base approach outperform their baseline and hand-constructed counterparts on all data availability points. A significant margin is maintained between 95\% confidence intervals. 
It is important to consider time requirements as well. In theory, the time requirement for the base approach is \(O(nk)\), where \(n\) is the number of modalities and \(k\) is a function showing the time taken to train the base model. However, all models used to take measurements are scaled down by \(1/n\), thus the time commitment in practice is closer to that of training the base CNN. However, if the topology were to be constructed by hand, either through a grid search or through trial and error, it would take more than a single try to arrive at a peak performing topology. Similarly, the number of hyper-parameters is reduced from \(O(n^2)\) of all pair-wise connections to only two. 
\begin{table}[t]
\renewcommand{\tabcolsep}{1.5mm}
\begin{center}
\caption{Comparison of accuracies of KerasNet based models}
\label{tab:res_keras}
\resizebox{\textwidth}{!}{
\begin{tabular}{l r r r r r}
\multicolumn{6}{c}{CIFAR-10}\\
\toprule
\(_{\small\mbox{Model}}\backslash^{p\%}\) & 20\% (\%) & 40\%  (\%) & 60\%  (\%) & 80\%  (\%) & 100\%  (\%)\\
\midrule
KerasNet & \(70.02\pm0.14\) & \(76.57\pm0.16\) & \(79.28\pm0.17\) & \(81.40\pm0.10\) & \(82.55\pm0.11\) \\
XKerasNet & \(71.00\pm0.23\) & \(76.92\pm0.10\) & \(79.62\pm0.16\) & \(81.32\pm0.10\) & \(82.68\pm0.15\) \\
Xsertion & \(\mathbf{72.02\pm0.70}\) & \(\mathbf{77.29\pm0.16}\) & \(\mathbf{79.92\pm0.07}\) & \(\mathbf{81.54\pm0.10}\) & \(\mathbf{82.93\pm0.09}\) \\ 
\bottomrule
\multicolumn{6}{c}{~}\\
\multicolumn{6}{c}{CIFAR-100}\\
\toprule
\(_{\small\mbox{Model}}\backslash^{p\%}\) & 20\% (\%) & 40\%  (\%) & 60\%  (\%) & 80\%  (\%) & 100\%  (\%)\\
\midrule
KerasNet & \(28.20\pm0.13\) & \(36.28\pm0.24\) & \(42.14\pm0.56\) & \(45.40\pm0.33\) & \(48.53\pm0.35\) \\
XKerasNet & \(30.41\pm0.32\) & \(39.32\pm0.39\) & \(43.95\pm0.40\) & \(47.08\pm0.23\) & \(48.96\pm0.22\) \\
Xsertion & \(\mathbf{31.31\pm0.49}\) & \(\mathbf{40.01\pm0.11}\) & \(\mathbf{44.74\pm0.20}\) & \(\mathbf{47.75\pm0.27}\) & \(\mathbf{50.29\pm0.53}\) \\
\bottomrule
\end{tabular}
}
\end{center}
\end{table}

\begin{table}[t]
\renewcommand{\tabcolsep}{1.5mm}
\begin{center}
\caption{Comparison of accuracies of FitNet based models}
\label{tab:res_fitnet}
\resizebox{\textwidth}{!}{
\begin{tabular}{l r r r r r}
\multicolumn{6}{c}{CIFAR-10}\\
\toprule
\(_{\small\mbox{Model}}\backslash^{p\%}\) & 20\% (\%) & 40\%  (\%) & 60\%  (\%) & 80\%  (\%) & 100\%  (\%)\\
\midrule
FitNet & \(75.47\pm0.32\) & \(82.02\pm0.18\) & \(84.98\pm0.20\) & \(86.22\pm0.19\) & \(87.42\pm0.05\) \\
XFitNet & \(76.56\pm0.24\) & \(82.43\pm0.07\) & \(85.11\pm0.19\) & \(86.23\pm0.18\) & \(87.42\pm0.08\) \\
Xsertion & \(\mathbf{77.35\pm0.15}\) & \(\mathbf{82.66\pm0.09}\) & \(\mathbf{85.43\pm0.12}\) & \(\mathbf{86.78\pm0.16}\) & \(\mathbf{87.77\pm0.22}\) \\ 
\bottomrule
\multicolumn{6}{c}{~}\\
\multicolumn{6}{c}{CIFAR-100}\\
\toprule
\(_{\small\mbox{Model}}\backslash^{p\%}\) & 20\% (\%) & 40\%  (\%) & 60\%  (\%) & 80\%  (\%) & 100\%  (\%)\\
\midrule
FitNet & \(29.29\pm1.69\) & \(40.91\pm2.48\) & \(50.94\pm0.51\) & \(55.47\pm0.96\) & \(58.92\pm0.60\) \\
XFitNet & \(36.17\pm0.27\) & \(48.02\pm0.72\) & \(54.18\pm0.36\) & \(57.98\pm0.33\) & \(60.32\pm0.29\) \\
Xsertion & \(\mathbf{38.59\pm0.37}\) & \(\mathbf{50.11\pm0.30}\) & \(\mathbf{55.48\pm0.41}\) & \(\mathbf{59.06\pm0.63}\) & \(\mathbf{61.67\pm0.31}\) \\
\bottomrule
\end{tabular}
}
\end{center}
\end{table}

\subsubsection{Iterative Approach.}
The iterative approach was further applied to optimise the automatically constructed networks. 15 iterations were used, training each model for a maximum of 30 epochs. 
On CIFAR-100 with KerasNet base CNN, the accuracy was observed to jump to \(51.07\%\) roughly \(0.3\%\) above the upper 95\% confidence bar. Similarly, on CIFAR-10, the accuracy jumped to \(83.36\%\), again roughly \(0.3\%\) above the upper 95\% confidence bar. In the base approach, luminance modality Y was deemed significantly more important than V, which in turn was slightly more important than U. The iterative approach strengthened Y\(\rightsquigarrow\)U connection and weakened U\(\rightsquigarrow\)Y. However, the converse was true for Y\(\rightsquigarrow\)V. It seems that it is important to transfer information from Y to U and from V to Y. Whilst connections between U and V did not disappear, they became significantly weakened both ways. For FitNet on CIFAR-10/100, the iterative approach did not yield any significant improvements. The learned connection weights were very close to those of the base approach.

\subsubsection{Application to Residual Learning.}
The base approach was further applied to networks utilising residual learning~\cite{he2016deep}, to see how it performs with state-of-art architectures. A variant of residual in residual network~\cite{targ2016resnet} was used here, which utilised pre-activations and contained 12 residual blocks. It was trained for 200 epochs, with a batch size of 64, using Adam optimiser~\cite{kingma2014adam}, with learning rate schedule of \(10^{-3},10^{-4},10^{-5}\) transitioning at 50 and 75 epochs. The layers were initialised as in~\cite{he2015delving} with \(10^{-4}\) \(L_2\) regularisation applied to all parameters.
The base network achieved 85.72\% on CIFAR-10 and 55.43\% on CIFAR-100 datasets. When the base approach was applied to architecture, with \(\alpha=2, \beta=2\), 80 epochs, the final networks achieved 88.81\% and 61.33\% on CIFAR-10/100, respectively, showing value and validity of methodology even for the latest networks.

\section{Conclusion}
We presented an way to apply the ideas of cross-modality and a library that helps to facilitate that. The results show that the automatically constructed topologies perform better than the baseline networks and hand-optimised X-CNNs, without adding additional parameters and reducing the time required to produce such topologies. The base approach successfully introduces difficult modification of CNN architecture using a data-driven procedure, which removes vast amounts of hyper-parameters that need to be considered. In that respect, the presented work reduces the complexity of applying a state-of-the-art advance in CNN design to a library call. Ideas behind combined learning procedure can be transferred to other work focused on building networks automatically from scratch, speeding up traversal of the vast search space. Hopefully, this shows that the bleeding-edge results in deep learning need not to exist in a vacuum. The ideas behind the work here can be transferred and applied to other research, resulting in a more automated field, which invites further adoption.

\bibliographystyle{splncs03_unsrt}
\bibliography{karazija2018autoxcnn}

\end{document}